\documentclass{article}

\usepackage{microtype}
\usepackage{graphicx}
\usepackage{subcaption}
\usepackage{booktabs} % for professional tables

\usepackage[accepted]{icml2019}

\usepackage{times}  %Required
\usepackage{helvet}  %Required
\usepackage{courier}  %Required
\usepackage{url}  %Required
\usepackage{amsfonts}
\usepackage{verbatim}
\usepackage{amsmath,amssymb}
\usepackage{nicefrac} 
\usepackage{multirow,tabulary,subcaption,wrapfig}

\icmltitlerunning{Batch Virtual Adversarial Training for Graph Convolutional Networks}

\begin{document}

\twocolumn[
\icmltitle{Batch Virtual Adversarial Training for Graph Convolutional Networks}




\begin{icmlauthorlist}
\icmlauthor{Zhijie Deng}{ts}
\icmlauthor{Yinpeng Dong}{ts}
\icmlauthor{Jun Zhu}{ts}
\end{icmlauthorlist}

\icmlaffiliation{ts}{Dept. of Comp. Sci. \& Tech., Institute for AI, BNRist Center, Tsinghua University, Beijing, 100084, China}

\icmlcorrespondingauthor{Jun Zhu}{dcszj@mail.tsinghua.edu.cn}

\icmlkeywords{Machine Learning, ICML}

\vskip 0.3in
]



\printAffiliationsAndNotice{}  % leave blank if no need to mention equal contribution

\begin{abstract}
We present batch virtual adversarial training (BVAT), a novel regularization method for graph convolutional networks (GCNs). BVAT addresses the shortcoming of GCNs that do not consider the smoothness of the model's output distribution against local perturbations around the input. We propose two algorithms, sample-based BVAT and optimization-based BVAT, which are suitable to promote the smoothness of GCN classifiers by generating virtual adversarial perturbations for either a subset of nodes far from each other or all nodes with an optimization process. Extensive experiments on three citation network datasets \emph{Cora}, \emph{Citeseer} and \emph{Pubmed} and a knowledge graph dataset \emph{Nell} validate the effectiveness of the proposed method, which establishes state-of-the-art results in the semi-supervised node classification task.
\end{abstract}

\section{Introduction}
\label{sec:inc}
\begin{figure*}[t]
  \centering
    \includegraphics[width=0.8\linewidth]{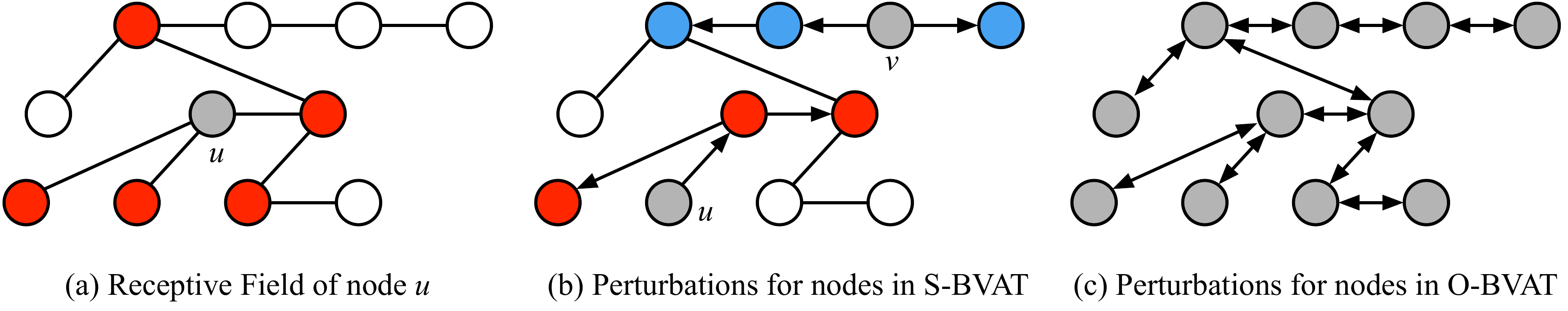}
      %\captionsetup{font={footnotesize}}
      \vspace{-2ex}
    \caption{(a) The receptive field (marked by red) of a node $u$ in two-layer GCNs. (b) In S-BVAT, two nodes $u$ and $v$ are selected to calculate the LDS loss, and the virtual adversarial perturbations are applied to the features in their receptive fields (marked by red and blue), which do not have intersection. (c) In O-BVAT, all nodes are included to calculate the LDS loss and the virtual adversarial perturbations for all nodes are optimized together.}
    \label{fig:framework}
    \vspace{-2ex}
\end{figure*}

Recent neural network models for graph-structured data \cite{kipf2017semi,hamilton2017inductive,velickovic2018graph} demonstrate remarkable performance in the semi-supervised node classification task. These methods essentially adopt different aggregators to aggregate feature information from the neighborhood of a node to obtain node prediction. The aggregators promote the smoothness between nodes in a neighborhood, which is helpful for semi-supervised node classification based on the assumption that connected nodes in the graph are likely to have similar representations \cite{kipf2017semi}.
%Graph convolutional networks (GCNs) \cite{kipf2017semi} extend powerful convolutional neural networks (CNNs) \cite{lecun1995convolutional} that deal with grid-structured data to model graph-structured data, which is presented by a set of nodes and edges. 
%GCNs can learn node representations by integrating the connectivity patterns and node features, thus enabling information propagation through edges. By conditioning the neural network model on the adjacency matrix of the graph, the gradient information can be distributed from nodes with labels to nodes without labels, and therefore the model can be applied to semi-supervised node classification.
However, these methods only consider the smoothness between nodes in a neighborhood without considering the smoothness of the output distribution of the node classifier. Previous works have confirmed that smoothing the output distribution of a classifier (i.e., encouraging the classifier to produce similar outputs) against local perturbations around the input can improve its generalization performance in supervised and especially semi-supervised learning \cite{wager2013dropout,sajjadi2016regularization,Laine2017Temporal,miyato2017virtual,luo2017smooth}.
Moreover, it's crucial to encourage the smoothness of the output distribution of aggregator-based graph models since the receptive field (e.g., Fig.~\ref{fig:framework}a) of a single node grows exponentially with respect to the number of aggregators in the model \cite{chen2017stochastic}, and neural network models tend to be non-smooth with such high dimensional input space \cite{goodfellow2014explaining,peck2017lower}. Therefore, it is necessary to encourage the \textbf{smoothness of the output distribution} of existing graph models.

Virtual adversarial training (VAT) \cite{miyato2017virtual, miyato2016adversarial} is an effective regularization method to encourage the smoothness of the output distribution of the classifier. %Instead of training the model to be isotropically smooth around each input, VAT defines the local distributional smoothness (LDS) at each input data point as the robustness of the model against the worst-case virtual adversarial perturbation.
%The virtual adversarial perturbation is approximated by the first dominant eigenvector of the Hessian matrix of the LDS loss.
%The local smoothness of the output distribution (virtual adversarial loss) is defined as the robustness of the model's posterior distribution against worst-case perturbation around a data point.
%VAT has demonstrated competitive performance in semi-supervised image recognition \cite{miyato2017virtual} and text classification tasks \cite{miyato2016adversarial}.
However, the straightforward extension of VAT to graph-based classification models is less effective.
%We notice that GCNs are trained by a batch gradient descent algorithm, where all nodes in the graph are processed together for forward and backward propagation.
%So we expect to deploy VAT to calculate the LDS loss for all nodes at the same time instead of only one node in each training iteration, for the purpose of efficient training.
%so that the whole algorithm is unbiased and the additional computation complexity is small.
The reason is that graph-based classifiers predict a node $u$ based on the features of all nodes in its receptive field (i.e., $RF_u$), as shown in Fig.~\ref{fig:framework}a, and consequently, the gradient of $u$'s loss can be back propagated to all nodes in $RF_u$. 
It means that the virtual adversarial perturbation calculated for node $u$ will modify the features of every node in $RF_u$.
Thus, when applying VAT into graph models and training with batch gradient descent \cite{kipf2017semi} or stochastic gradient descent \cite{hamilton2017inductive,chen2017stochastic}, once the batch (or mini-batch) contains other nodes whose receptive fields include $u$, the virtual adversarial perturbations generated for $u$ and those nodes overlap. 
%all those nodes will have overlap with the perturbations for node $u$. 
As a result, the overall perturbation for node $u$ is actually not the worst-case virtual adversarial perturbation, making VAT inefficient to encourage the smoothness of the model's output distribution and unable to push decision boundaries of the model away from real data instances effectively. %There are also some works concentrating on applying mini-batch training into GCNs \cite{hamilton2017inductive,chen2017stochastic}, but the mini-batch training has the same problem as the whole batch training because of the possible nodes' interaction within each batch.
%so the proposed algorithms are suitable for these works as well.

Given the aforementioned issue, we aim to generate virtual adversarial perturbations \textbf{perceiving the connectivity patterns between nodes in the graph} to promote the smoothness of the node classifier's output distribution. %so that the LDS loss really reflects the worst-case smoothness at every nodes. 
%Based on this motivation, 
In this paper, we propose batch virtual adversarial training (BVAT) algorithms. % in this paper, which are appropriate for GCNs to promote the smoothness of the output distribution at all nodes.
%of the node classifier and ensure the robustness of the classifier against local perturbations around the receptive field of each node. 
Specifically, we focus on the typical and effective graph convolutional networks (GCNs) \cite{kipf2017semi}, and propose a sample-based BVAT algorithm (S-BVAT) to craft local virtual adversarial perturbations for a subset of separable nodes and an optimization-based BVAT algorithm (O-BVAT) that generates adversarial perturbations at all nodes.
BVAT exploits the high efficiency of batch gradient descent in GCNs. To validate the effectiveness of BVAT, we conduct experiments on four challenging node classification benchmarks: \emph{Cora}, \emph{Citeseer}, \emph{Pubmed} citation datasets as well as a knowledge graph dataset \emph{Nell}. BVAT establishes state-of-the-art results across all datasets with a tolerable additional computation complexity. %We also make extensive comparisons between BVAT and VAT for graph-structured data and show the superiority of BVAT.

\section{Related Work}

Learning node representations based on graph for semi-supervised learning and unsupervised learning has drawn increasing attention and has been developed mainly toward two directions: spectral approaches \cite{zhu2003semi,belkin2006manifold,weston2012deep,defferrard2016convolutional,kipf2017semi} and non-spectral approaches \cite{perozzi2014deepwalk,tang2015line,grover2016node2vec,yang2016revisiting,monti2017geometric,hamilton2017inductive,velickovic2018graph}. %On one hand, label propagation \cite{zhu2003semi}, manifold regularization \cite{belkin2006manifold}, deep semi-supervised embedding \cite{weston2012deep}, Chebyshev expansion based spatially localized filters \cite{defferrard2016convolutional} and graph convolutional networks \cite{kipf2017semi} inherit the ideas from spectral graph theory \cite{ekambaram2013wavelet} and demonstrate impressive results in the context of node classification. On the other hand, non-spectral approaches learn graph embeddings directly on spatially local neighborhoods. DeepWalk \cite{perozzi2014deepwalk} and its variants \cite{tang2015line,grover2016node2vec} learn node representations based on the neighborhood generated by random walks. Planetoid \cite{yang2016revisiting}, MoNet \cite{monti2017geometric} and GraphSAGE \cite{hamilton2017inductive} propose end-to-end frameworks for learning embeddings for semi-supervised learning or unsupervised learning on graph. Recently, graph attention networks \cite{velickovic2018graph} introduce masked self-attentional layers into graph convolutions and establish a strong baseline for transductive and inductive learning on graph.
There is also an interest in applying regularization terms \cite{miyato2017virtual,Laine2017Temporal,tarvainen2017mean,luo2017smooth} to semi-supervised learning based on the cluster assumption \cite{chapelle2005semi}.  
%Various sophisticated solutions have been proposed \cite{miyato2017virtual,Laine2017Temporal,tarvainen2017mean,luo2017smooth}, which achieve striking results. 
Among them, virtual adversarial training (VAT) has been proved successful in various domains \cite{miyato2016adversarial,miyato2017virtual}. However, VAT is not effective enough when straightforwardly applied to the models that deal with graph-structured data because of the interrelationship between different nodes, as stated in Sec.~\ref{sec:inc}. Thus we propose a novel regularization BVAT to address this issue. The works of adversarial attacks on graph-structure data \cite{zugner2018adversarial,dai2018adversarial} also share the idea of considering the connectivity patterns of the graph to generate adversarial perturbations, but our work focuses more on semi-supervised node classification instead of performing adversarial attacks. Moreover, graph partition neural networks proposed by \cite{liao2018graph} has the similar idea of splitting original graph into disjoint subgraphs to S-BVAT, but it is dedicated to promoting the efficiency of information propagation in graph rather than the smoothness of the output distribution. %in the paper which introduces a novel regularization term to smooth the output distribution of the models.

\section{Batch Virtual Adversarial Training}

In this section, we first extend virtual adversarial training (VAT) \cite{miyato2017virtual} to graph convolutional networks (GCNs) and discuss its shortcomings. We then propose the batch virtual adversarial training (BVAT) algorithms which are more suitable for GCNs. %In particular, we introduce the sample-based batch virtual adversarial training (S-BVAT) and the optimization-based batch virtual adversarial training (O-BVAT) respectively. 
\subsection{Virtual Adversarial Training for Graph Convolutional Networks}

\label{sec:VAT}
Virtual adversarial training (VAT) \cite{miyato2017virtual} encourages the  smoothness by training the model to be robust against local worst-case virtual adversarial perturbation.
In VAT, the local distributional smoothness (LDS) is defined by a virtual adversarial loss as
\begin{equation}
\label{eq:3}
\mathrm{LDS}(x, \mathcal{W}, r_{\mathrm{vadv}}) = D_{\mathrm{KL}}\big(p(y|x, \hat{\mathcal{W}})||p(y|x+r_{\mathrm{vadv}}, \mathcal{W})\big),
\end{equation}
where $p(y|x, \mathcal{W})$ is the prediction distribution parameterized by $\mathcal{W}$ (i.e., trainable parameters), $D_{\mathrm{KL}}(\cdot,\cdot)$ is the KL divergence of two distributions, $\hat{\mathcal{W}}$ denotes the current estimation of the parameters $\mathcal{W}$ and $r_{\mathrm{vadv}}$ is the virtual adversarial perturbation found by
\begin{equation}
\label{eq:4}
\begin{split}
r_{\mathrm{vadv}}&=\mathop{\arg\max}\limits_{r; ||r||_2 \leq \epsilon}\mathrm{LDS}(x, \mathcal{W}, r) \\
&=\mathop{\arg\max}\limits_{r; ||r||_2 \leq \epsilon}D_{\mathrm{KL}}\big(p(y|x, \hat{\mathcal{W}})||p(y|x+r, \mathcal{W})\big).
\end{split}
\end{equation}
% The closed form solution for $r_{\mathrm{vadv}}$ does not exist. In \cite{miyato2017virtual}, $\mathrm{LDS}(x, \mathcal{W}, r)$ is approximated by the second-order Taylor expansion and $r_{\mathrm{vadv}}$ emerges as the first dominant eigenvector of the Hessian matrix of $\mathrm{LDS}(x, \mathcal{W}, r)$ with respect to $r$.
\begin{comment}
When VAT is applied to semi-supervised classification, the full objective function to minimize is given by
\begin{equation}
\label{eq:5}
\mathcal{L} = \frac{1}{|D_l|}\sum\limits_{x_l,y_l \in D_l}\ell(x_l, y_l, \mathcal{W}) + \alpha\cdot\frac{1}{|D|}\sum\limits_{x \in D} \mathrm{LDS}(x, \mathcal{W}, r_{\mathrm{vadv}}),
\end{equation}
where the first term is the supervised loss at labeled data points and the second term is the virtual adversarial loss at all data points.
\end{comment}
%VAT has demonstrated impressive results in semi-supervised image and text classification \cite{miyato2016adversarial,miyato2017virtual}.

%When applying VAT to GCNs for semi-supervised node classification tasks, 
A straightforward extension into GCNs 
%of the LDS loss at each node is defined as $\mathrm{LDS}(X_u, \mathcal{W}, r_{\mathrm{vadv},u})$, where $X_u$ is the input feature matrix of all nodes in $u$'s receptive field $R_u$. In training GCNs, we use 
is using the average LDS loss for all nodes as a regularization term
\begin{equation}
\label{eq:(3)}
\mathcal{R}_{\mathrm{vadv}}(\mathcal{V}, \mathcal{W}) = \frac{1}{N}\sum\limits_{u \in \mathcal{V}} \mathrm{LDS}(X_u, \mathcal{W}, r_{\mathrm{vadv},u}),
\end{equation}
where $\mathcal{V}$ denotes the node set of the graph containing $N$ elements and $X_u$ is the input feature matrix of all nodes in $RF_u$. $r_{\mathrm{vadv},u}$ is the virtual adversarial perturbation matrix for node $u$ in the same size as $X_u$ and is approximated by the first dominant eigenvector of the Hessian matrix of $\mathrm{LDS}(X_u, \mathcal{W}, r)$ 
%, we approximate $\mathrm{LDS}(X_u, \mathcal{W}, r)$ with the second-order Taylor expansion and find the first dominant eigenvector of the Hessian matrix of $\mathrm{LDS}(X_u, \mathcal{W}, r)$ with respect to $r$ as $r_{\mathrm{vadv}, u}$ 
using a power iteration method with $T$ iterations \cite{miyato2017virtual}. The overall loss is
\begin{equation}
\label{eq:6}
\begin{split}
\mathcal{L} = \mathcal{L}_0 + \alpha \cdot\frac{1}{|\mathcal{V}|}\sum\limits_{u \in \mathcal{V}}E\big(p(y|X_u, \mathcal{W})\big) + \beta\cdot \mathcal{R}_{\mathrm{vadv}}(\mathcal{V}, \mathcal{W}),
\end{split}
\end{equation}
where $\mathcal{L}_0$ is the average cross-entropy loss of all labeled nodes and $E(\cdot)$ is the conditional entropy of a distribution, 
%The conditional entropy of the output distribution for all nodes in $\mathcal{V}$ serves as an additional regularization term to encourage one-hot predictions, 
which is widely used in the semi-supervised classification task \cite{grandvalet2005semi} to encourage one-hot predictions. $\alpha$ and $\beta$ are coefficients for conditional entropy and local distributional smoothness. %$\mathcal{L}_0$ is the average cross-entropy loss of nodes in the labeled nodes set $\mathcal{V_L}$:
% \begin{equation}
% \label{eq:2}
% \mathcal{L}_0 = \frac{1}{|\mathcal{V_L}|}\sum\limits_{u \in \mathcal{V_L}}\ell\big(y_u, p(y|X_u, \mathcal{W})\big).
% \end{equation}

Notably, the interaction of nodes in the graph reduces the effectiveness of loss $\mathrm{LDS}(X_u, \mathcal{W},r_{\mathrm{vadv},u})$ since in the batch training, the calculated $r_{\mathrm{vadv},u}$ is the accumulation of the virtual adversarial perturbations generated for all nodes in $RF_u$. Therefore, $\|r_{\mathrm{vadv},u}\|_2\leq \epsilon$ in Eq.~\eqref{eq:4} cannot be guaranteed and  the resultant perturbations are \textbf{not} the worst-case virtual adversarial perturbations.
%we notice that a major difference between VAT on graph-structured data and other forms of data (e.g., images, texts) is that the prediction of one node relies on the input features of others, indicating that the virtual adversarial perturbation generated for one node will modify the features of other nodes in its receptive field. 
%And if we generate virtual adversarial perturbations for all nodes in a batch or mini-batch for efficient training, the perturbation applied to a node is actually the accumulation of the virtual adversarial perturbations generated by many nodes, in whose receptive field this node lies. So the resultant perturbations are \textbf{not} the worst-case virtual adversarial perturbation (see Sec.~\ref{sec:efficiency}). Therefore, the virtual adversarial loss $\mathrm{LDS}(X_u, \mathcal{W},r_{\mathrm{vadv},u})$ may not reflect the robustness of the model's output distribution and cannot effectively encourage the local smoothness at each input instance.

\subsection{Batch Virtual Adversarial Training}

%In this section and the following section, we elaborate the proposed batch virtual adversarial training (BVAT) algorithms including the sample-based BVAT (S-BVAT) and the optimization-based BVAT (O-BVAT).
BVAT can perceive the connectivity patterns between nodes and alleviate the interaction effect of virtual adversarial perturbations crafted for all nodes by either stochastically sampling a subset of nodes far from each other or adopting a more powerful optimization process for generating virtual adversarial perturbations. These two approaches are both harmonious with the batch gradient descent optimization method used by GCNs and only increase tolerable additional computation complexity, as shown in Appendix~\ref{sec:time}. 
%When applying virtual adversarial training to graph models, we should remain the interactions between connected nodes so that the information can be passed through connections and the nodes in a neighborhood have similar representations.
%We also expect to reduce the influence of one node's worst-case adversarial perturbations on the receptive field of the other nodes which are applied VAT at the same time. Based on the motivations, we propose batch virtual adversarial training(BVAT) algorithms, including BVAT\_sample and BVAT\_opt, and we are going to describe them in detail.

\textbf{S-BVAT}. The motivation of sample-based BVAT is that we expect to make the model be aware of the relationship between nodes and limit the propagation of adversarial perturbations to prevent perturbations from different nodes interacting with each other. In S-BVAT, we generate virtual adversarial perturbations for a subset $\mathcal{V_S} \subset \mathcal{V}$ of nodes, whose receptive fields do not overlap with each other.
%The receptive field of a node in GCN grows exponentially with the respect to the number of graph convolutional layers because there is an aggregation operation in every layer.
Taking a $K$-layer GCN model for example, the receptive field $RF_u$ of a node $u$ contains all the $k$-hop neighbors of it where $0 \leq k \leq K$. If we expect $RF_u$ doesn't have intersection with $RF_v$, the number of nodes in the shortest path between $u$ and $v$ (denoted as the distance $D_{uv}$) should be at least $2K$ (shown in Fig.~\ref{fig:framework}b). 
Therefore, we randomly sample a subset $\mathcal{V_S}$ of nodes with a fixed size $B$ (e.g., 100) as
\begin{equation*}
\mathcal{V_S} = \{ u | u \in \mathcal{V} \}, \;\; \text{s.t.} \; |\mathcal{V_S}| = B, \; \forall u,v \in \mathcal{V_S}, D_{uv} \geq 2K.
\end{equation*}
%where each node is far away from the others so that any two nodes have no overlapping of receptive fields and local worst-case virtual adversarial perturbations.
In this way, the generated perturbations for nodes in $\mathcal{V_S}$ do not interact with each other.
%Thus, we can deploy VAT for every node in $\mathcal{V_S}$ at the same time based on the batch gradient descent method:
The regularization term for training is the average LDS loss over nodes in $\mathcal{V_S}$ as
\begin{multline}
\label{eq:8}
\mathcal{R}_{\mathrm{vadv}}(\mathcal{V_S}, \mathcal{W}) = \frac{1}{B}\sum\limits_{u \in \mathcal{V_S}}LDS(X_u,\mathcal{W},r_{\mathrm{vadv},u}).
\end{multline}
$\mathcal{R}_{\mathrm{vadv}}(\mathcal{V_S}, \mathcal{W})$ can be seen as an approximate estimation of $\mathcal{R}_{\mathrm{vadv}}(\mathcal{V}, \mathcal{W})$. %$r_{\mathrm{vadv},u}$ is the matrix of virtual adversarial perturbation in the same size as $X_u$, and is applied to the nodes in $u$'s receptive field. $r_{\mathrm{vadv},u}$ can be found by Eq.~\ref{eq:4}. 
% \begin{equation}
% \begin{split}
% \label{eq:7}
% r_{\mathrm{vadv}, u}&=\mathop{\arg\max}\limits_{r; ||r||_F \leq \epsilon} \mathrm{LDS}(X_u, \mathcal{W}, r)\\
% &= D_{\mathrm{KL}}\big(p(y|X_u, \hat{\mathcal{W}})||p(y|X_u+r, \mathcal{W})\big).
% \end{split}
% \end{equation}
%Similar to VAT  \cite{miyato2017virtual}, we approximate $\mathrm{LDS}(X_u, \mathcal{W}, r)$ with the second-order Taylor expansion and find the first dominant eigenvector of the Hessian matrix of $\mathrm{LDS}(X_u, \mathcal{W}, r)$ with respect to $r$ as $r_{\mathrm{vadv}, u}$ by a power iteration method \cite{golub2001eigenvalue} with $T$ iterations. Note that 
The virtual adversarial perturbations for all nodes in $\mathcal{V_S}$ can be processed at the same time in the batch gradient descent. %And that's why we call the proposed algorithms batch virtual adversarial training. We use $R$ to denote the perturbations of all nodes in the graph, which has the same size as $X$.
% The overall objective function for training semi-supervised node classification GCNs is
% \begin{equation}
% \label{eq:9}
% \mathcal{L} = \mathcal{L}_0 + \alpha \cdot\frac{1}{|\mathcal{V}|}\sum\limits_{u \in \mathcal{V}}E\big(p(y|X_u, \mathcal{W})\big) + \beta\cdot \mathcal{R}_{\mathrm{vadv}}(\mathcal{V_S}, \mathcal{W}),
% \end{equation}
% where $\mathcal{L}_0$ is the supervised loss defined in Eq.~\eqref{eq:2} and $E(\cdot)$ is the conditional entropy of a distribution. $\alpha$ and $\beta$ are hyper-parameters to balance the loss terms.
As suggested by \cite{miyato2017virtual} and our experiments, one-step power iteration is sufficient for approximating $r_{\mathrm{vadv}, u}$ and obtaining high performance. We summarize S-BVAT in Algorithm~\ref{algo:1} in Appendix.%So we simply use $T=1$ in algorithm~\ref{algo:1} and we further discuss the effect of $T$ in Sec.~\ref{sec:efficiency}.

% \begin{algorithm}[t] 
% \caption{Sample-based batch virtual adversarial training (S-BVAT)} 
% \begin{algorithmic}[1]
% \State $\mathcal{V_S}=\varnothing$, $\mathcal{V_C}=\mathcal{V}$.
% \While {$|\mathcal{V_S}| < B$}
% \State Choose a node $u$ from $\mathcal{V_C}$ randomly and add $u$ to $\mathcal{V_S}$.
% \State Remove all nodes in the $k$-hop ($\forall k \in [0, 2K]$) neighborhood of $u$ from $\mathcal{V_C}$.
% \State Initialize $r_{\mathrm{vadv}, u}$ from an iid Gaussian distribution and normalize it as $\|r_{\mathrm{vadv}, u}\|_F=1$.
% \EndWhile 
% \State Calculate $r_{\mathrm{vadv}, u}$ by taking the gradient of $\mathrm{LDS}(X_u, \mathcal{W}, r)$ with respect to $r$:
% $$g_{u} \leftarrow \nabla_r D_{\mathrm{KL}}\big(p(y|X_u, \hat{\mathcal{W}})||p(y|X_u+r, \mathcal{W})\big) |_{r=\xi r_{\mathrm{vadv}, u}},$$
% $$r_{\mathrm{vadv}, u} = \epsilon \cdot g_u / \|g_u\|_F.$$
% \State \textbf{return}
% $\nabla_\mathcal{W} \mathcal{R}_{\mathrm{vadv}}(\mathcal{V_S}, \mathcal{W}) |_{\mathcal{W}=\hat{\mathcal{W}}}.$
% \end{algorithmic} 
% \label{algo:1} 
% \end{algorithm}

%On the other hand, we want to find universal worst-case virtual adversarial perturbations $R_{vadv,\mathcal{V}}$ which can be applied on $X$ so that the receptive field $R_u$ of every node $u$ are attacked in the worst-case direction. 

\textbf{O-BVAT}. %In the perspective of batch training, Eq.~\eqref{eq:4} of VAT for GCN can be formalized as
% \begin{equation*}
% \mathop{\arg\max}\limits_{R; ||R_v||_2 \leq \epsilon, v \in \mathcal{V}} \frac{1}{N}\sum\limits_{u \in \mathcal{V}}D_{\mathrm{KL}}\big(p(y|X_u, \hat{\mathcal{W}})||p(y|X_u+R_u, \mathcal{W})\big),
% \end{equation*}
% where $R$ is the adversarial perturbations matrix for the whole feature matrix $X$ and $R_u$ denotes some rows of $R$ that related to the prediction of node $u$.
% The VAT algorithm in Sec.~\ref{sec:VAT} search for $R$ using the second-order Taylor approximation of the average LDS loss but it has two problems: 1) the limitation $||R_v||_2 \leq \epsilon$ cannot be ensured; 2) $R_v$ for each node $v$ are not in the worst-case adversarial direction. These problems are caused by the interaction between the LDS loss at nearby nodes as discussed in Sec.~\ref{sec:VAT}.
%
%To overcome the problems of VAT for GCN, in the optimization-based batch virtual adversarial training (O-BVAT) algorithm, 
In an alternative way, we propose to generate virtual adversarial perturbations for all nodes in $\mathcal{V}$ by an optimization process, which proves to be more powerful in adversarial attacks than one-step gradient-based methods \cite{carlini2017towards}. We maximize the average LDS loss with respect to the whole perturbation matrix $R$ corresponding to the whole feature matrix $X$ so that the neighborhood perturbations $R_u$ (i.e., $r_{\mathrm{vadv},u}$) of every node $u$ are adversarial enough. At the same time, we punish the norm of $R$ so that the perturbations are small enough.
%It's a difficult problem as $\mathcal{G}$ contains many connections and they form a complicated information passing architecture. We cannot use the original virtual adversarial training algorithm to calculate $R_{vadv,\mathcal{V}}$ because the perturbations obtained by original VAT are not in the worst-case direction as we talk about in Sec.\ref{sec:VAT}. As a result, we draw inspiration from FGSM which is widely used in adversarial sample domain and can produce adversarial perturbations efficiently, and we propose an optimization based batch adversarial training algorithm BVAT\_opt. BVAT\_opt first searches for universal worst-case perturbations by:

Specifically, $R$ is optimized by solving
\begin{multline}
\label{eq:10}
\mathop{\max}\limits_{R} \frac{1}{N}\sum\limits_{u \in \mathcal{V}}D_{\mathrm{KL}}\big(p(y|X_u, \hat{\mathcal{W}})||p(y|X_u+R_u, \mathcal{W})\big) \\ - \gamma \cdot \|R\|_F^2,
%=\frac{1}{|\mathcal{V}|}\sum\limits_{u \in \mathcal{V}}KLD(p(y_u|X, \hat{W})||p(y_u|X+R, \hat{W})) - \gamma * norm(R)
\end{multline}
where $\|R\|_F$ is the Frobenius norm of $R$ which makes the optimal perturbation have a small norm, and $\gamma$ is a hyper-parameter to balance the loss terms. 
We optimize $R$ with an Adam \cite{KingmaB14} optimizer for $T$ iterations.
The regularization term in O-BVAT is then the average LDS loss over all nodes in $\mathcal{V}$, similar to Eq.~\eqref{eq:(3)}.
We summarize O-BVAT in Algorithm~\ref{algo:2} in Appendix.
%except that the regularization term is the average LDS loss over the entire node set $\mathcal{V}$ as $\mathcal{R}_{\mathrm{vadv}}(\mathcal{V}, \mathcal{W})$.
%\begin{equation}
%\label{eq:12}
%O(X,W) = \frac{1}{|\mathcal{V_L}|}\sum\limits_{v \in \mathcal{V_L}}CE(y_v, h_v^{(K)})+\alpha * \frac{1}{|\mathcal{V}|}\sum\limits_{v \in \mathcal{V}}entropy(h_v^{(K)}) + \beta* LDS(X, W, \mathcal{V})
%\end{equation}
%where
%\begin{equation}
%\label{eq:11}
%LDS(X, W, \mathcal{V}) = \frac{1}{|\mathcal{V}|}\sum\limits_{u \in \mathcal{V}}KLD(p(y_u|X, \hat{W})||p(y_u|X+R_{vadv, \mathcal{V}}, W)).
%\end{equation}

\section{Experiments}

We empirically evaluate the BVAT algorithms through experiments on different datasets. %We first specify the experimental settings in Sec.~\ref{sec:setup}.
%We then demonstrate the effectiveness of the BVAT algorithms by showing that the regularization term can well represent the worst-case local smoothness and the models trained by our methods are quite smooth against perturbations in Sec.~\ref{sec:efficiency}.
Owing to promoting the smoothness of the model's output distribution, the BVAT algorithms boost the performance of GCNs significantly and achieve superior results on four popular benchmarks against a wide variety of state-of-the-art methods, which is detailed in Sec.~\ref{sec:semi}.
%Moreover, we test the effects of hyper-parameters on the performance of BVAT in Sec.~\ref{sec:eff-sample} and we talk about the additional computation complexity of BVAT in Sec.~\ref{sec:time}.
We implement BVAT based on the official implementation of GCNs and the experimental setup is detailed in Appendix~\ref{sec:setup}.

%and utilize a different early stopping strategy on \emph{Pubmed} and \emph{Nell}: we report the test accuracy of the model at the epoch when the model achieve highest validation accuracy. 
\begin{figure*}[ht]
\centering
\begin{subfigure}{0.3\linewidth}
  \centering
  \includegraphics[width=\linewidth]{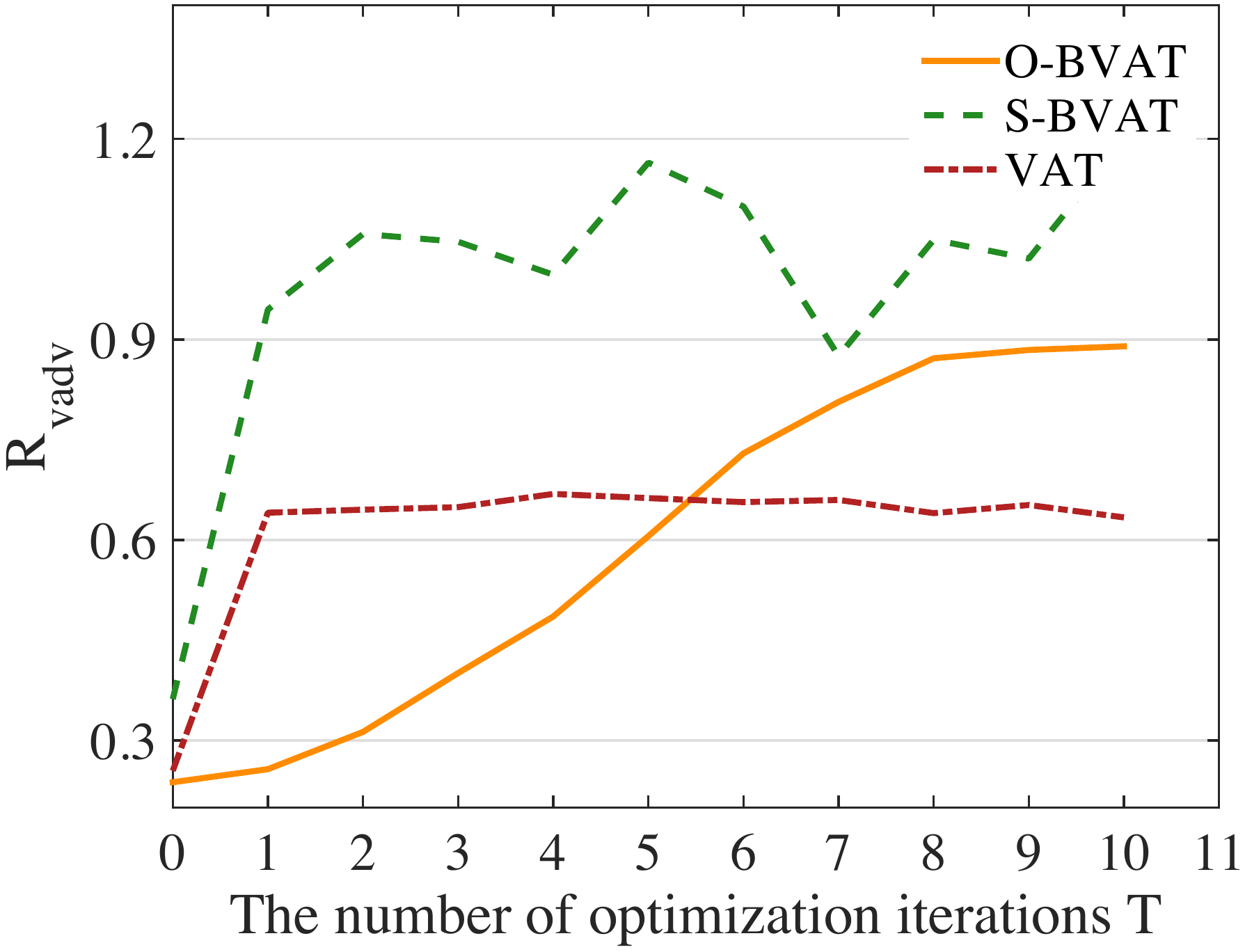}
  \caption{}
  \label{fig:1}
\end{subfigure}
\begin{subfigure}{0.3\linewidth}
  \centering
  \includegraphics[width=\linewidth]{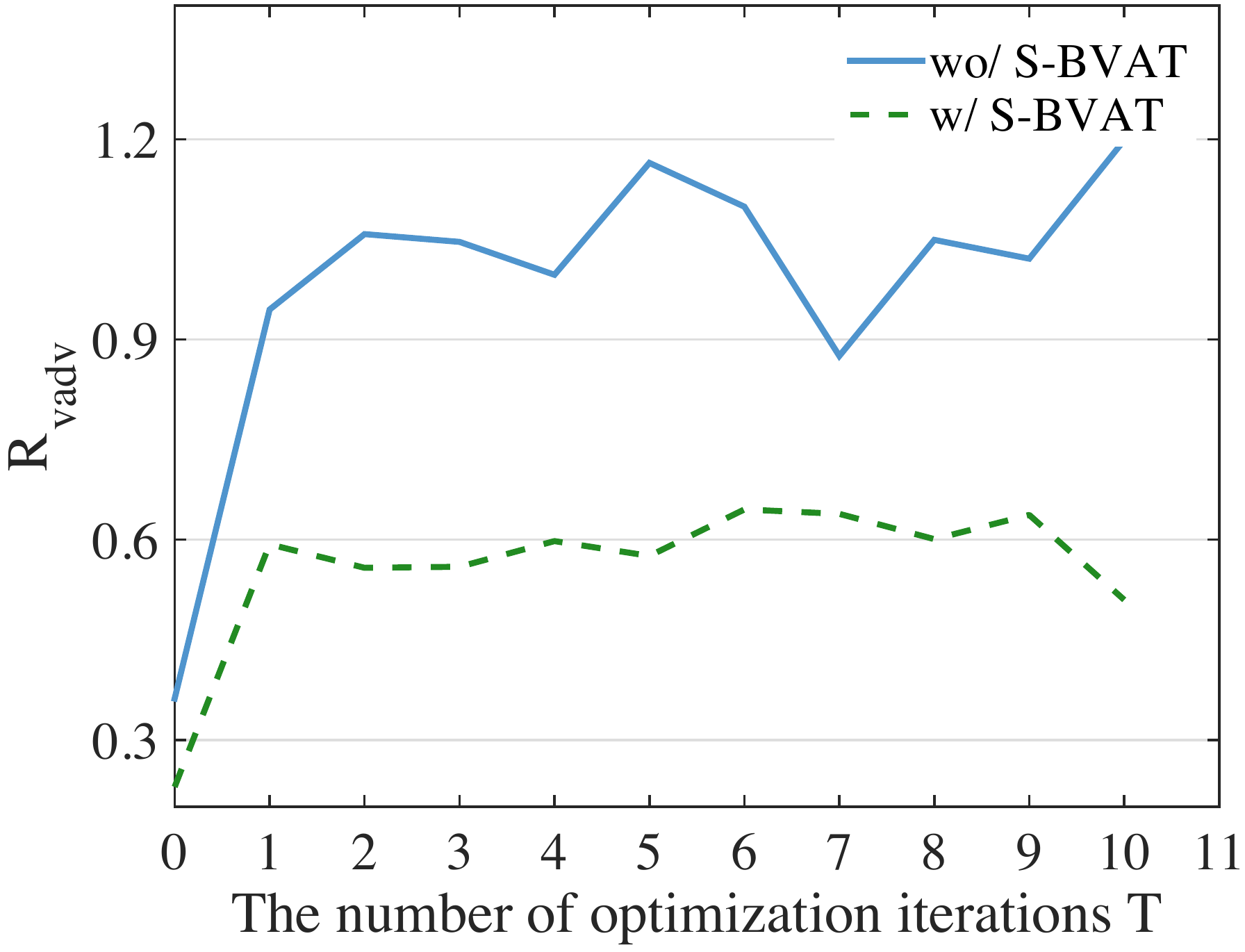}
  \caption{}
  \label{fig:2}
\end{subfigure}
\begin{subfigure}{0.3\linewidth}
  \centering
  \includegraphics[width=\linewidth]{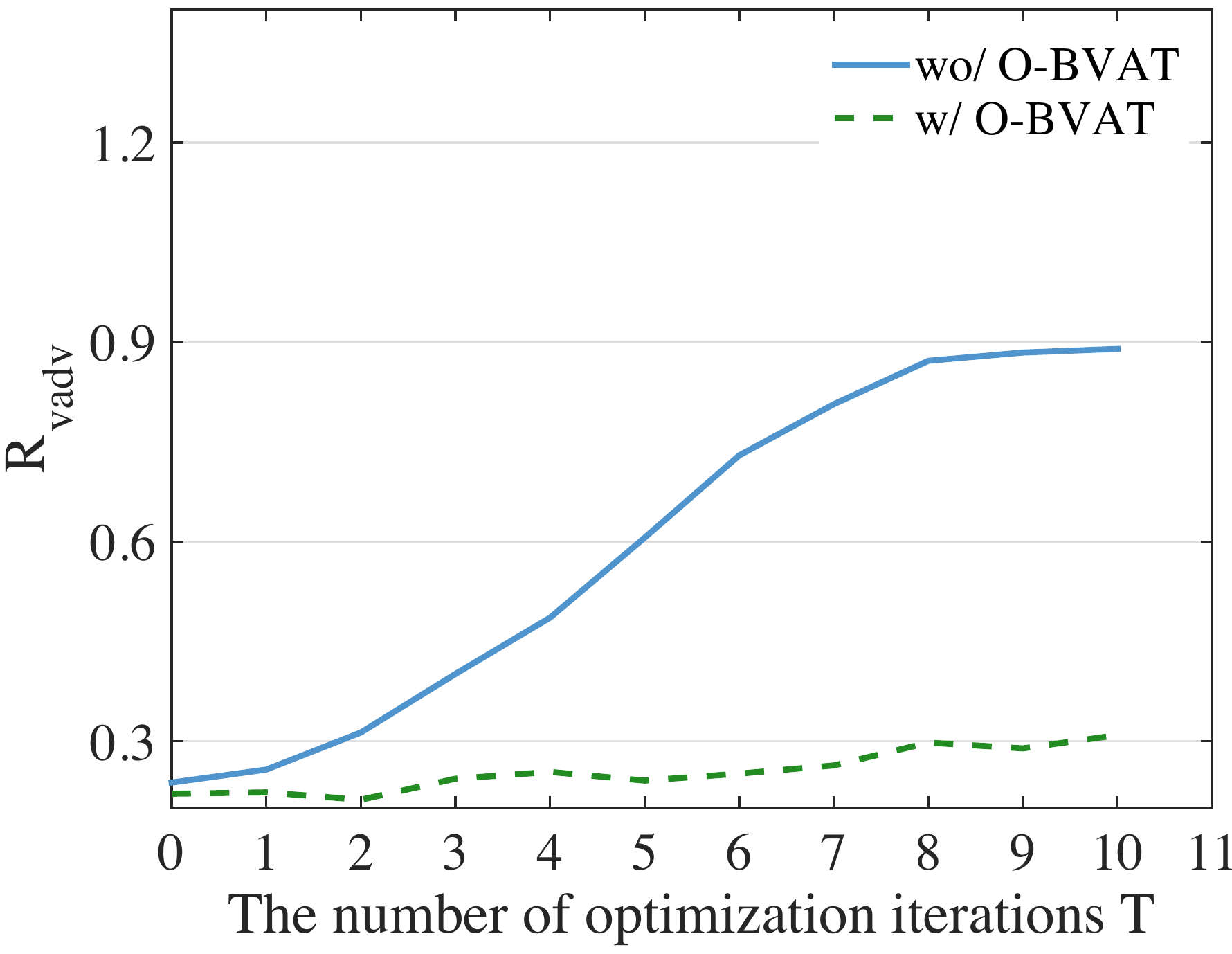}
  \caption{}
  \label{fig:2-1}
\end{subfigure}
\vspace{-1ex}
\caption{(a) Comparisons of $\mathcal{R}_{\mathrm{vadv}}$ of VAT, S-BVAT and O-BVAT on a baseline GCN model. (b) Comparisons of $\mathcal{R}_{\mathrm{vadv}}$ of S-BVAT on models trained with and without S-BVAT. (c) Comparisons of $\mathcal{R}_{\mathrm{vadv}}$ of O-BVAT on models trained with and without O-BVAT.}
\vspace{-2ex}
\end{figure*}

\begin{table*}[t]
  \caption{Summary of node classification results in terms of test accuracy (\%).}
  \label{table:2}  
  \centering
  \begin{tabular*}{1.0\textwidth}{@{\extracolsep{\fill}}lllll}
    \toprule
    \textbf{Method}     & \textbf{Cora}     & \textbf{Cireseer} & \textbf{Pubmed} & \textbf{Nell} \\
    \midrule
    ManiReg \cite{belkin2006manifold}  & 59.5 & 60.1  &  70.7 & 21.8 \\
    SemiEmb \cite{weston2012deep}     & 59.0 & 59.6 & 71.1 & 26.7     \\
    LP \cite{zhu2003semi}     & 68.0    & 45.3 & 63.0 & 26.5  \\
    DeepWalk \cite{perozzi2014deepwalk} &67.2& 43.2 &65.3&58.1\\
    Planetoid \cite{yang2016revisiting} &75.7 & 64.7 & 77.2 & 61.9\\
    Monet \cite{monti2017geometric} &81.7 $\pm$ 0.5 & -- & 78.8 $\pm$ 0.3 & --\\
    GAT \cite{velickovic2018graph} &83.0 $\pm$ 0.7 & 72.5 $\pm$ 0.7 & 79.0 $\pm$ 0.3 & --\\
    GPNN \cite{liao2018graph} &81.8 & 69.7 & 79.3 & 63.9\\
    \midrule
    GCN \cite{kipf2017semi} &81.5 & 70.3 & 79.0 & 66.0\\
    GCN w/ random perturbations & 82.3 $\pm$ 2.0 & 71.4 $\pm$ 1.9 & 79.2 $\pm$ 0.6 & 65.9 $\pm$ 1.0\\
    \midrule
    \textbf{GCN w/ VAT} & 82.8 $\pm$ 0.8 & 73.0 $\pm$ 0.7 & 79.5 $\pm$ 0.3 & 66.0 $\pm$ 1.1 \\
    \textbf{GCN w/ S-BVAT} &83.4 $\pm$ 0.6 & 73.1 $\pm$ 1.3 & 79.6 $\pm$ 0.5 & 66.0 $\pm$ 0.9\\
    \textbf{GCN w/ O-BVAT} &\textbf{83.6} $\pm$ 0.5 & \textbf{74.0} $\pm$ 0.6 & \textbf{79.9} $\pm$ 0.4 & \textbf{67.1} $\pm$ 0.6\\
    \bottomrule
  \end{tabular*}
  \vspace{-1ex}
\end{table*}

\subsection{Effectiveness of BVAT}

\label{sec:efficiency}
We evaluate the effectiveness of the BVAT algorithms by assessing the virtual adversarial perturbations generated by them for graph data. 
First, we train a vanilla GCN model on \emph{Cora}. 
Then, we use VAT, S-BVAT and O-BVAT to manufacture virtual adversarial perturbations and calculate the regularization term $\mathcal{R}_{\mathrm{vadv}}$ averaged on all the nodes $\mathcal{V}$ (in VAT and O-BVAT) or a subset of nodes $\mathcal{V_S}$ (in S-BVAT). $\mathcal{R}_{\mathrm{vadv}}$ indicates whether the perturbations are worst-case locally adversarial or not.
We plot $\mathcal{R}_{\mathrm{vadv}}$ in Fig.~\ref{fig:1}.

It is clear that O-BVAT and S-BVAT achieve higher $\mathcal{R}_{\mathrm{vadv}}$ values than VAT, which demonstrates that BVAT can find virtual adversarial perturbations which are more likely to be in the worst-case direction. % and more effective and helpful for virtual adversarial training. 
For both S-BVAT and VAT, We can observe a significant jump of the $\mathcal{R}_{\mathrm{vadv}}$ value when $T$ changes from 0 (random perturbations) to 1, but no much more with larger $T$. Therefore we set $T=1$ in VAT and S-BVAT. However, $\mathcal{R}_{\mathrm{vadv}}$ of O-BVAT increases monotonically with respect to $T$ and converges at about 10 steps, so we set $T=10$ in the following experiments.

%o-BVAT 0.23771831 0.25722027 0.31297502 0.40102598 0.4856351 0.6058415 0.6896519 0.7466495 0.821875 0.88431287 0.8997711
%S-BVAT 0.36200345, 0.9447379, 1.0577955, 1.04618, 0.9968915, 1.1644835, 1.0988963, 0.8752998, 1.0491891, 1.0210016, 1.1985886
%VAT 0.25486952, 0.6409105, 0.6454126, 0.64919215, 0.6689437, 0.6627189, 0.6567123, 0.65996027, 0.6402562, 0.65240175, 0.6338513

On the other hand, we compare the robustness (i.e., smoothness in the worst-case direction) of the models trained with S-BVAT/O-BVAT with that of the models trained without S-BVAT/O-BVAT by calculating the $\mathcal{R}_{\mathrm{vadv}}$ values.
Fig.~\ref{fig:2} and~\ref{fig:2-1} show the results respectively. The models trained with BVAT (S-BVAT or O-BVAT) have lower $\mathcal{R}_{\mathrm{vadv}}$ values than the vanilla GCN models, which indicates that the models trained with BVAT are more robust against the  adversarial perturbations and more smooth in the input space.
%The smoothness of the model's output distribution will in consequence improve the generalization performance, as showed in the next section.

%s-bvat
%without 0.36200345, 0.9447379, 1.0577955, 1.04618, 0.9968915, 1.1644835, 1.0988963, 0.8752998, 1.0491891, 1.0210016, 1.1985886
%with 0.22968352, 0.59309876, 0.55794466, 0.5594623, 0.59791255, 0.575755, 0.6453744, 0.6387284, 0.6008774, 0.637164, 0.509938
%o-bvat
%without  0.36281827 0.37082893 0.39559865 0.50013006 0.58541626 0.68767416 0.8481217 0.9969723 1.1752993 1.3984599
%with  0.27090174 0.2729753 0.2618744 0.29351452 0.3037139 0.29068875 0.3011411 0.31344888 0.3477793 0.33913743

\subsection{Semi-supervised Node Classification}
\label{sec:semi}
%BVAT can decrease model's $\mathcal{R}_{\mathrm{vadv}}$ effectively and improve model's smoothness and robustness against local perturbations. Therefore, BVAT should have improved generalization performance on node classification tasks. 
To empirically validate the effectiveness of smoothing output distribution, we deploy BVAT and VAT algorithms for semi-supervised node classification on the \emph{Cora}, \emph{Citeseer}, \emph{Pubmed} and \emph{Nell}, and compare with state-of-the-art methods in Table~\ref{table:2}. 
We also train a GCN model with random input perturbations as a baseline.
We report the averaged results of 10 runs with different random seeds.

% To keep comparisons fair, we adopt the same network architecture, initialization and hyper-parameters as GCNs and report the averaged results of 10 runs with different random seeds. $\pm$ in Table 2 represents the standard deviation of test accuracy of 10 runs.

The proposed GCNs with VAT, GCNs with S-BVAT and GCNs with O-BVAT all outperform the vanilla GCNs and GCNs trained with random input perturbations by a large margin across all the four datasets. Furthermore, as expected, O-BVAT boosts the performance significantly and establishes state-of-the-art results. %When comparing the performance between S-BVAT and O-BVAT, we find that O-BVAT demonstrates better performance because the 
O-BVAT uses the LDS loss on all nodes, which may be more efficient than that on a subset of nodes used by S-BVAT. %We also notice that the performance of GCNs trained with VAT is only a little worse than GCNs trained with S-BVAT. The reason is that though the perturbations produced by GCNs trained with VAT are not in the worst-case direction, they are adversarial and could help to improve smoothness to some extent. In one iteration, GCNs with VAT make use of all the nodes in the graph which are more than B nodes used by GCNs with S-BVAT to do virtual adversarial training, so it can explore the input space more sufficiently to meet the clustering assumption of classifier. 

\begin{comment}
\subsection{Computation Complexity Analysis}
\label{sec:time}
%Actually BVAT algorithms will only bring a tolerable additional computation complexity because BVAT algorithms work in a batch manner and they only need to calculate the gradients of the LDS loss with respect to the input feature matrix without updating the parameters of the graph convolutional neural networks classifier. 
We empirically estimate the time consuming of GCN, GCN with S-BVAT and GCN with O-BVAT on \emph{Cora} dataset, and they averagely need 0.0355154, 0.03946125 and 0.1064431 seconds for one epoch on a GTX 1080Ti respectively. GCN with S-BVAT is a little slower than GCN as there are only two additional forward propagations and one additional back propagation. GCN with O-BVAT spends less than $3\times$ time than GCN because the optimization process involves T + 1 additional forward propagations and T additional back propagations (T = 10 in all the experiments). %Considering that the classification performance of GCN W/ BVAT is noticeably better than GCN, it’s no doubt that the extra calculation cost is acceptable.
\end{comment}

\section{Conclusion}
In this paper, we proposed batch virtual adversarial training algorithms, which can smooth the output distribution of graph-based classifiers and are essentially suitable for any aggregator-based graph neural networks. In particular, we presented sample-based batch virtual adversarial training and optimization-based virtual adversarial training algorithms respectively. %, which address the issues of vanilla virtual adversarial training method applied to graph-structured data. 
Experimental results demonstrate the effectiveness of the proposed method on various datasets in the semi-supervised node classification task. BVAT outperforms the current state-of-the-art methods by a large margin. 

\bibliography{example_paper}
\bibliographystyle{icml2019}

\clearpage
\appendix

\section{Algorithms for BVAT}
We present the detailed algorithms for sample-based batch virtual adversarial training (S-BVAT) and optimization-based batch virtual adversarial training (O-BVAT) in Algorithm~\ref{algo:1} and Algorithm~\ref{algo:2}, respectively.

\begin{algorithm}[h]
   \caption{Sample-based batch virtual adversarial training (S-BVAT)}
   \label{algo:1}
\begin{algorithmic}[1]
   \STATE $\mathcal{V_S}=\varnothing$, $\mathcal{V_C}=\mathcal{V}$.
    \WHILE {$|\mathcal{V_S}| < B$}
    \STATE Choose a node $u$ from $\mathcal{V_C}$ randomly and add $u$ to $\mathcal{V_S}$.
    \STATE Remove all nodes in the $k$-hop ($\forall k \in [0, 2K]$) neighborhood of $u$ from $\mathcal{V_C}$.
    \STATE Initialize $r_{\mathrm{vadv}, u}$ from an iid Gaussian distribution and normalize it as $\|r_{\mathrm{vadv}, u}\|_F=1$.
    \ENDWHILE 
    \STATE Calculate $r_{\mathrm{vadv}, u}$ by taking the gradient of $\mathrm{LDS}(X_u, \mathcal{W}, r)$ with respect to $r$:
    $$g_{u} \leftarrow \nabla_r D_{\mathrm{KL}}\big(p(y|X_u, \hat{\mathcal{W}})||p(y|X_u+r, \mathcal{W})\big) |_{r=\xi r_{\mathrm{vadv}, u}},$$
    $$r_{\mathrm{vadv}, u} = \epsilon \cdot g_u / \|g_u\|_F.$$
    \STATE \textbf{return}
    $\nabla_\mathcal{W} \mathcal{R}_{\mathrm{vadv}}(\mathcal{V_S}, \mathcal{W}) |_{\mathcal{W}=\hat{\mathcal{W}}}.$
\end{algorithmic}
\end{algorithm}

\begin{algorithm}[h] 
\caption{Optimization-based batch virtual adversarial training (O-BVAT)} 
\label{algo:2}
\begin{algorithmic}[1]
\STATE Initialize $R^{(0)} \in \mathbb{R}^{N\times D}$ from an iid Gaussian distribution.
\FOR{$i$ = 1 to $T$} 
\STATE Calculate the gradient of Eq.~\eqref{eq:10} with respect to $R^{(i-1)}$ as $g^{(i)}$.
%$$G \leftarrow - \nabla_R L(X,R,{\mathcal{V}}) - \gamma * norm(R) |_{R= R^{(i-1)}} .$$
\STATE Use an Adam optimizer to perform gradient ascent as $R^{(i)} \leftarrow \textrm{Adam}(R^{(i-1)}, g^{(i)})$.
%$$R^{(i)} \leftarrow adam(R^{(i-1)}, G).$$ 
\ENDFOR 
\STATE $R \leftarrow R^{(T)}$.
\STATE \textbf{return} $\nabla_{\mathcal{W}} \mathcal{R}_{\mathrm{vadv}}(\mathcal{V}, \mathcal{W}) |_{\mathcal{W}=\hat{\mathcal{W}}}$.
\end{algorithmic} 
\end{algorithm}

\section{Experimental Setup} 
\label{sec:setup}
We examine BVAT on the three citation network datasets \emph{Cora}, \emph{Citeseer} and \emph{Pubmed} \cite{sen2008collective} and one knowledge graph dataset \emph{Nell} \cite{yang2016revisiting} with the same train/validation/test splits as \cite{yang2016revisiting} and \cite{kipf2017semi}. The details of the four datasets are summarized in Table~\ref{t-1}. 
We use the same preprocessing strategies as GCNs. The dimension of the preprocessed node features in \emph{Nell} is $61,278$, so the input sparse matrix $X \in \mathbb{R}^{65755 \times 61278}$ is too large to be converted to a dense matrix that doesn't exceed the GPU memory (GTX 1080Ti). As a result, BVAT algorithms construct sparse virtual adversarial perturbations $R$ for \emph{Nell}. We use the same architecture, initialization, dropout rate, $L_2$ regularization factor, number of hidden units and number of epochs as GCNs.
%Considering this, we implement the two BVAT algorithms that are suitable for processing sparse tensors and constructing sparse virtual adversarial perturbations.
\begin{table}[t]
  \caption{Statistics of the datasets used in our experiments.}
  \label{t-1}
  \centering
  \begin{tabular*}{0.45\textwidth}{@{\extracolsep{\fill}}lrrrr}
    \toprule
         & \textbf{Cora}     & \textbf{Cireseer} & \textbf{Pubmed} & \textbf{Nell} \\
    \midrule
    \textbf{Nodes}  & 2,708 & 3,327  &  19,717 & 65,755 \\
    \textbf{Edges}     & 5,429 & 4,732 & 44,338 & 266,144     \\
    \textbf{Features}     & 1,433       & 3,703 & 500 & 61,278  \\
    \textbf{Classes} &7& 6 &3&105\\
    \textbf{Label rate} &0.052 & 0.036 & 0.003 & 0.001\\
    \bottomrule
  \end{tabular*}
\end{table}
%\textbf{Configurations.}
%We utilize the same architecture and initialization as the original 2-layer GCNs and keep the hyper-parameters of GCNs (dropout rate for two layers, $L_2$ regularization factor for the first layer, number of hidden units and number of epochs) unchanged. 
For S-BVAT, we fix $B=100$, $\xi=10^{-6}$ and $T=1$. We set the perturbation size $\epsilon=0.03$ for \emph{Cora} and \emph{Citeseer} and $\epsilon=0.003$ for \emph{Pubmed} and \emph{Nell}. For O-BVAT, we run an Adam optimizer with learning rate $0.001$ for $T=10$ iterations. We set $\gamma=0.01$ on \emph{Pubmed} and $\gamma=1$ on the others. We tune the hyper-parameters $\alpha$ and $\beta$ because the label rate, feature size and number of edges vary significantly across different datasets.

\begin{figure*}[h]
\centering
\begin{subfigure}{0.45\textwidth}
  \centering
  \includegraphics[width=1.0\linewidth]{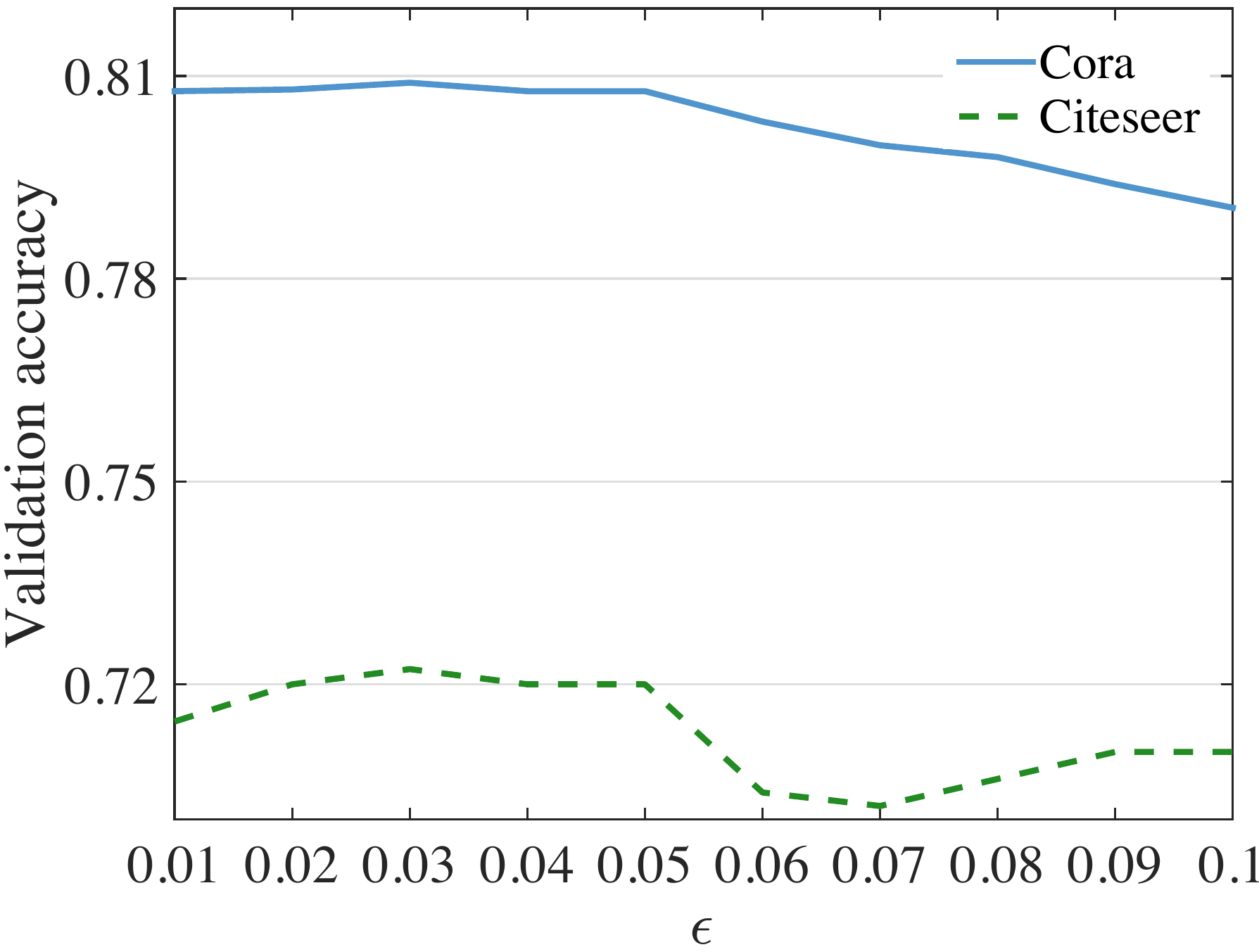}
  \caption{}
  \label{fig:3}
\end{subfigure}
\hspace{5ex}
\begin{subfigure}{0.45\textwidth}
  \centering
  \includegraphics[width=1.0\linewidth]{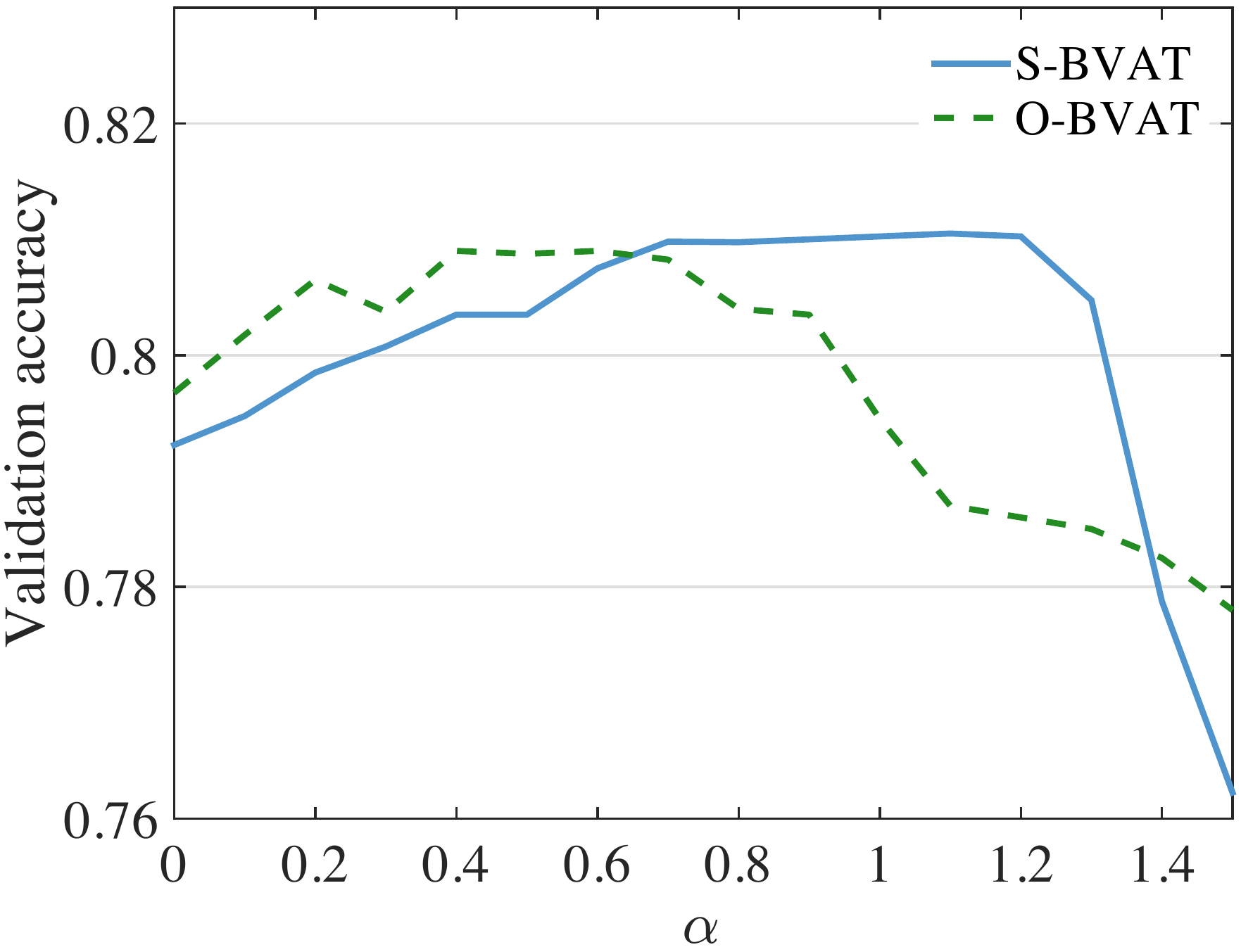}
  \caption{}
  \label{fig:4}
\end{subfigure}
\caption{(a) Effect of $\epsilon$ on the validation performance of S-BVAT for \emph{Cora} and \emph{Citeseer}. (b) Effect of $\alpha$ on the validation performance of S-BVAT and O-BVAT for \emph{Cora}.}
\end{figure*}

\section{Ablation study of $\epsilon$ and $\alpha$}
\label{sec:eff-sample}
To investigate how the perturbation size $\epsilon$ in S-BVAT affects the final classification results, we conduct an ablation study in Fig.~\ref{fig:3}, where we plot the validation accuracy of the trained models on \emph{Cora} and \emph{Citeseer} with respect to the varied $\epsilon$ while keeping the other hyper-parameters fixed (on \emph{Cora}, we set $\beta=1.2$ and $\alpha=0.7$; on \emph{Citeseer}, we set $\beta=0.8$ and $\alpha=0.7$). As we have observed, S-BVAT is not sensitive to $\epsilon$ when it changes from 0.01 to 0.1 and we choose $\epsilon=0.03$ for both \emph{Cora} and \emph{Citeseer}. The conclusion is also true for \emph{Pubmed} and \emph{Nell}, where $\epsilon$ is set to $0.003$ due to the smaller norm of input features in these two datasets.

% 0.01 - 0.1 
% 0.80775 0.808 0.809 0.8077500000000001 0.80775 0.80325 0.79975 0.798 0.794 0.7905000000000001
% 

% 0.7144999999999999 0.72 0.7222499999999999 0.72 0.72 0.704 0.702 0.706 0.71 0.71
Incorporating the conditional entropy term $E(\cdot)$ into training of semi-supervised tasks is confirmed useful generally \cite{grandvalet2005semi}. We expect to determine the importance of the role it plays in the BVAT algorithms. Thus, we conduct two set of experiments on \emph{Cora} by assessing the validation accuracy of the models trained with different values of $\alpha$ in the range $[0, 1.5]$ with a granularity $0.1$. The results of the two algorithms (we assign $\beta$ to 1.2 and 1.5 for S-BVAT and O-BVAT respectively) are plotted in Fig.~\ref{fig:4}. The results demonstrate that S-BVAT and O-BVAT can remain high performance with regularization coefficient $\alpha$ varying in a large range. Therefore, we think that the virtual adversarial perturbations used in BVAT play a crucial role in smoothing the output distribution of the model and improving its performance, while the conditional entropy term $E(\cdot)$ is an extra regularization which made the model more suitable for semi-supervised classification.
% 0.79225 0.7947500000000001 0.7985 0.8007500000000001 0.8035000000000001 0.8035000000000001 0.8075000000000001 0.809 0.8097500000000001 0.81 0.8102499999999999 0.8105 0.81025 0.8047500000000001 0.77875 0.76225
% 0.7967500000000001 0.8017500000000001 0.8065 0.8037500000000001 0.8089999999999999 0.8067500000000001 0.809 0.80825 0.804 0.8035 0.7945000000000001 0.787 0.786 0.785 0.7825 0.778

\section{Computation Complexity Analysis}
\label{sec:time}
Actually BVAT algorithms will only bring a tolerable additional computation complexity because BVAT algorithms work in a batch manner and they only need to calculate the gradients of the LDS loss with respect to the input feature matrix without updating the parameters of the graph convolutional neural networks classifier. 
We empirically estimate the time consuming of GCN, GCN with S-BVAT and GCN with O-BVAT on \emph{Cora} dataset, and they averagely need 0.0355154, 0.03946125 and 0.1064431 seconds for one epoch on a GTX 1080Ti respectively. GCN with S-BVAT is a little slower than GCN as there are only two additional forward propagations and one additional back propagation. GCN with O-BVAT spends less than $3\times$ time than GCN because the optimization process involves $T + 1$ additional forward propagations and $T$ additional back propagations ($T = 10$ in all the experiments). Considering that the classification performance of GCN w/ BVAT is noticeably better than GCN, it’s no doubt that the extra calculation cost is acceptable.


\end{document}